\documentclass[10pt,conference]{IEEEtran}

\usepackage{cite}
\ifCLASSINFOpdf
   \usepackage[pdftex]{graphicx}
   \graphicspath{{figs/}}
   \DeclareGraphicsExtensions{.pdf,.jpeg,.png}
\else
   \usepackage[dvips]{graphicx}
   \graphicspath{{../figs/}}
   \DeclareGraphicsExtensions{.eps}
\fi
\usepackage[cmex10]{amsmath}
\usepackage{amsthm}

\usepackage{algorithmic}
\usepackage{array}
\ifCLASSOPTIONcompsoc
  \usepackage[caption=false,font=normalsize,labelfont=sf,textfont=sf]{subfig}
\else
  \usepackage[caption=false,font=footnotesize]{subfig}
\fi
\usepackage{url}
\usepackage{booktabs}
\usepackage{subfig}
\usepackage{multirow}
\usepackage{float}
\usepackage{xcolor}

\newif\iffinal
\finaltrue
\newcommand{\cmtid}{192}

\iffinal
\else
\usepackage[switch]{lineno}
\fi

\newif\ifreview
\reviewfalse

\ifreview
\usepackage[normalem]{ulem}
\fi

\iffinal
\IEEEoverridecommandlockouts
\fi

\begin{document}
%
\thispagestyle{empty}
\onecolumn
\linespread{1.2}\selectfont{}
{\noindent\Huge IEEE Copyright Notice}\\[1pt]

{\noindent\large Copyright (c) 2026 IEEE

\noindent Personal use of this material is permitted. Permission from IEEE must be obtained for all other uses, in any current or future media, including reprinting/republishing this material for advertising or promotional purposes, creating new collective works, for resale or redistribution to servers or lists, or reuse of any copyrighted component of this work in other works.}\\[1em]

{\noindent\Large Accepted to be published in: 2026 39th SIBGRAPI Conference on Graphics, Patterns and Images (SIBGRAPI'26), September 29 -- October 2, 2026.}\\[1in]

{\noindent\large Cite as:}\\[1pt]

{\setlength{\fboxrule}{1pt}
 \fbox{\parbox{0.65\textwidth}{A. C. Portella, S. F. Santos, and J. Almeida, ``Unsupervised Transfer Clustering for Mitigating Cold Start in Active Prompt Learning'' in \emph{2026 39th SIBGRAPI Conference on Graphics, Patterns and Images (SIBGRAPI)}, Goiânia, GO, Brazil, 2026, pp. 1--6}}}\\[1in]
 
{\noindent\large BibTeX:}\\[1pt]

{\setlength{\fboxrule}{1pt}
 \fbox{\parbox{0.95\textwidth}{
 @InProceedings\{SIBGRAPI\_2026\_Portella,
 
 \begin{tabular}{lll}
  & author    & = \{A. C. \{Portella\} and 
                    S. F. \{Santos\} and 
                    J. \{Almeida\}\},\\
			   
  & title     & = \{Unsupervised Transfer Clustering for Mitigating Cold Start in Active \\
  &           & \ \ \ \ Prompt Learning\}, \\
			   
  & pages     & = \{1--6\},\\
  
  & booktitle & = \{2026 39th SIBGRAPI Conference on Graphics, Patterns and Images (SIBGRAPI)\},\\
  
  & address   & = \{Goiânia, GO, Brazil\},\\
  
  & month     & = \{September 29 -- October 2\},\\
  
  & year      & = \{2026\},\\
  
  & publisher & = \{\{IEEE\}\},\\
  
  \end{tabular}
  
\}
 }}}

\twocolumn
\linespread{1}\selectfont{}
\clearpage

\title{Unsupervised Transfer Clustering for Mitigating Cold Start in Active Prompt Learning}


\iffinal

\author{
    \IEEEauthorblockN{
        André Camargo Portella,
        Samuel Felipe dos Santos, and
        Jurandy Almeida
    }
    \IEEEauthorblockA{
        Department of Computing, Federal University of S\~{a}o Carlos (UFSCar), Sorocaba, SP -- Brazil\\
        Emails: {\small\texttt{andre.portella@estudante.ufscar.br}} and {\small\texttt{\{samuel.felipe, jurandy.almeida\}@ufscar.br}}
    }
}

\else
  \author{SIBGRAPI Paper ID: \cmtid \\ }
  \linenumbers
\fi

\maketitle

\begin{abstract}

Vision-Language Models (VLMs) are able to achieve impressive zero-shot classification performance by aligning visual and textual representations, but each new task still demands handcrafted prompts.
Active Prompt Learning (APL) combines Active Learning (AL) and Prompt Learning (PL) into a single framework, allowing for the usage of the VLM prior knowledge for iteratively querying the most informative images to be labeled.  
However, the cold-start problem is still relevant for APL methods, where the performance of the initial query can be worse than random sampling.
While recent state-of-the-art APL methods mitigate with balanced sampling and multimodal features, they rely on rigid, distance-based clustering to group these features.
This simplistic approach can struggle to capture the complex, high-dimensional semantic distributions inherent to VLMs, leading to suboptimal query representativeness.
Unsupervised transfer can be applied to these features as a possible alternative, since it is capable of inferring the underlying human labeling of a task without any form of supervision. This way, samples can be grouped in semantically coherent clusters. 
This paper proposes Unsupervised Transfer Clustering with Selective Querying (UTC+SQ), a framework that enhances a recent APL approach by leveraging state-of-the-art unsupervised transfer model.
These models generate high-fidelity pseudo-labels that establish semantically meaningful clusters, allowing for the selection of more relevant samples.
Experimental evaluations demonstrate that shifting from distance-based to projection-based clustering improves the representativeness of the queried subset, achieving accuracy gains in 6 of the 8 datasets tested.

\end{abstract}

\IEEEpeerreviewmaketitle

\section{Introduction}
\label{sec:introduction}

Vision-Language Models (VLMs) have achieved state-of-the-art performance by aligning visual and textual information into a single representation. By pre-training on massive datasets, VLMs acquire rich, generalizable representations that enable impressive zero-shot classification capabilities across a wide array of downstream tasks without the need for task-specific training data, drastically reducing the dependency on large, exhaustively annotated datasets~\cite{kim2025active}.

Despite their impressive zero-shot performance, adapting VLMs to specific target domains still typically demands carefully handcrafted prompts, which can be time-consuming and labor-intensive~\cite{kim2025active}.
To address this, Prompt Learning (PL) methods can be used, allowing for parts of the prompt vector to be learned during training, but they usually require labeled target data~\cite{bang2024active}.
Active Learning (AL) strategies can be used to reduce annotation budget by iteratively querying the most informative sample to be labeled by an oracle.

However, AL methods can suffer from the cold-start problem, where the model fails to select samples that are more informative than a random baseline during the initial iterations\iffinal~\cite{saltori2022lowbudget}\fi.
Because AL is an iterative process, an unrepresentative initial query can severely impact the performance of all subsequent rounds~\cite{chen2022making}.
Active Prompt Learning (APL) frameworks emerge as a powerful solution that combines PL with AL.
In it, AL is used to iteratively query a human oracle to label only the most informative and representative samples. Then PL is used to learn the prompts for the downstream task.
This strategy maximizes the model's performance while keeping low human annotation budget~\cite{kim2025active}.

However, recent APL methods, such as Cluster-Balanced Acquisition with Selective Querying (CB+SQ)~\cite{kim2025active}, rely on simple distance-based clustering strategies for sample selection, such as the $k$-means algorithm. This simplistic approach is fundamentally limited, since it relies solely on spatial distances in the feature space and often ignores the complex, underlying semantic relationships of the data.

A possible alternative to distance-based clustering is unsupervised transfer learning. This paradigm assumes that only the number of classes is known, utilizing pre-trained representations of VLM models to solve a new task by uncovering the underlying human labeling entirely without supervision or task-specific representation learning~\cite{turtle}.

In this paper, we propose Unsupervised Transfer Clustering with Selective Querying (UTC+SQ), an APL framework that enhances the CB+SQ~\cite{kim2025active} approach by leveraging state-of-the-art unsupervised transfer models to improve the selection of examples. By replacing the $k$-means algorithm with a smarter clustering approach driven by unsupervised transfer, we establish a highly representative subset of images from the start.
Experimental evaluations demonstrate that shifting from distance-based to projection-based clustering improves the representativeness of the queried subset. Specifically, selecting the samples with the highest confidence within these semantic clusters yields superior prompt-tuning performance over the baseline, achieving accuracy gains in 6 of the 8 datasets tested.

The main contributions of this work are:

\begin{itemize}
\item We propose the Unsupervised Transfer Clustering with Selective Querying (UTC+SQ) framework\footnote{Our source code is available at \url{https://github.com/andre-portella/UTC}.}, that builds over a recent state-of-the-art APL method by generating semantically meaningful clusters.
\item Our strategy mitigates the cold-start problem by using high-fidelity unsupervised transfer pseudo-labels that more accurately represent the target domain from the start, improving the initial query of samples.
\item Our proposed UTC+SQ was able to obtain the best average accuracy over the 8 evaluated datasets, improving over the state-of-the-art baseline in 6 of them.
\end{itemize}

\section{Related Works}
\label{sec:related_works}

This section presents works related to ours, including VLM, PL, APL, and unsupervised transfer approaches. 

\subsection{Vision-Language Models and Prompt Learning}
\label{sec:VLM_prompt}

VLMs have achieved remarkable zero-shot performance across various tasks.
For instance, CLIP~\cite{radford2021learning} uses contrastive learning to map images and text into a shared embedding space, while GPT-4~\cite{achiam2023gpt} is a multimodal transformer capable of processing integrated image and text inputs.
Despite these advances, adapting these models to specific target domains still relies on handcrafted text prompts.

PL has emerged as a prominent alternative to adapt pre-trained models by directly optimizing learnable prompt vectors.
CoOp~\cite{zhou2022learning} represents prompts as learnable vectors while keeping pre-trained parameters frozen.
MaPLe~\cite{khattak2023maple} optimizes prompts across both vision and language branches to improve cross-modal alignment.
ProGrad~\cite{zhu2023prompt} introduces a principled, gradient-based constraint that prevents prompt tuning from forgetting the general knowledge.
Despite these improvements, PL techniques still require labeled data to guide optimization.

\subsection{Active Prompt Learning}
\label{sec:APL_refs}

Recent AL works leverage the prior knowledge of pre-trained VLMs with strong zero-shot capabilities to enhance budget efficiency~\cite{kim2025active}.
DropQuery focuses on low budget AL regimes by proposing an acquisition strategy that balances dropout-estimated uncertainty with representativeness~\cite{gupte2024revisiting}.
ActiveLLM utilizes large language models to autonomously select the most informative instances for training textual classifiers targeting the few-shot cold-start problem~\cite{bayer2026activellm}.

Combining PL and AL specifically for VLMs, some recent APL frameworks have been proposed.
The PCB~\cite{bang2024active} framework addresses the inherent class imbalance of standard AL by explicitly leveraging VLM priors to ensure a balanced sample selection.
CB+SQ~\cite{kim2025active} uses class-guided clustering, selecting the most representative sample of each cluster to be labeled. After the first round, adaptive class-wise confidence thresholds are used to assign pseudo-labels to high-confidence samples while querying annotators only for uncertain ones.
Despite the efficiency and state-of-the-art results presented by CB+SQ~\cite{kim2025active}, it currently relies on simple, distance-based clustering strategies to select samples, an approach that may ignore complex, underlying semantic relationships of the data.

\subsection{Unsupervised Transfer} 
\label{sec:unsupervised_transfer}

HUME~\cite{HUME} is a model-agnostic unsupervised transfer framework that infers human labeling by searching for consistent labelings across multiple fixed pre-trained representation spaces. HUME operates on the key insight that classes defined by human labeling are generally linearly separable. By seeking the labeling assignment that demonstrates high consistency across diverse encoders, HUME recovers an underlying labeling that is well-correlated with ground-truth data distributions~\cite{HUME}.
More recently, TURTLE~\cite{turtle} advanced over this framework by framing unsupervised transfer as a optimization problem that seeks a dataset labeling assignment that induces maximal-margin classifiers within the continuous representation spaces of pre-trained models.
By using VLMs to guide this search, TURTLE effectively uncovers the semantic structures of the target dataset and generates high-fidelity pseudo-labels that consistently outperform standard zero-shot transfer and unsupervised prompt-tuning baselines.

\section{Prompt Learning with Unsupervised Transfer Clustering}
\label{sec:our_method}

\begin{figure}[htbp]
    \centering
    \includegraphics[width=0.9\linewidth]{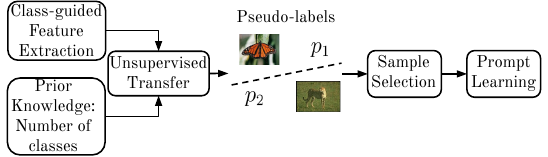}
    \caption{Overview of one APL round of our proposed UTC+SQ framework for mitigating the cold-start problem in APL. }
    \label{fig:overview}
\end{figure}

In this section, we describe our UTC+SQ framework for mitigating the cold-start problem in APL. Figure~\ref{fig:overview} presents an overview of our approach, where we keep the class-guided features from CB+SQ~\cite{kim2025active}, but we replace its simplistic distance-based clustering technique by state-of-the-art unsupervised transfer models that are able to generate semantically meaningful pseudo-labels using only these features and the number of classes.
Using these pseudo-labels we define clusters and select the most informative sample from each one. These samples are then annotated by a human oracle and used for prompt learning.
This process is repeated for each round of APL, where we kept the same annotation budget per round and budget-saving selective querying from CB+SQ~\cite{kim2025active}.

The remainder of this section describe in more detail each component of our framework.

\subsection{Problem Definition} 
\label{sec:problem_definition}

Let $\mathcal{D}_{U} = \{x_i\}_{i=1}^{N}$ be a pool of $N$ unlabeled images and $\mathcal{Y} = \{y_1, \dots, y_C\}$ be the class labels. A VLM with frozen pre-trained image $E_{I}$ and text $E_{T}$ encoders can be used for zero-shot predictions using a prompt $p=$\textit{``a photo of a}\ $[class]$\textit{''}, where $[class]$ represents the class name.

PL adapts the frozen VLM to downstream tasks by replacing handcrafted text queries with a set of continuous, learnable prompt vectors $V = \{v_1, \dots, v_M\}$ that are concatenated with the class identifier $[class]$. The probability of an image $x$ belonging to class $y$ is computed via the cosine similarity between the image feature and text feature.

While PL assumes a pre-existing labeled dataset, APL optimize the prompt vectors $V$ by selecting the most informative samples respecting a strict total annotation budget $\mathcal{B}$.

Starting with an empty labeled set $\mathcal{D}_{L} = \emptyset$, APL operates over $T$ iterations. At each iteration $t$, the framework executes the following steps:
(i) an acquisition function $\mathcal{A}$ evaluates the unlabeled pool $\mathcal{D}_{U}$ and selects a subset $\mathcal{S}_t \subset \mathcal{D}_{U}$ limited by a per-round budget $b$;
(ii) a human oracle provides the ground-truth labels for the selected subset $\mathcal{S}_t$;
(iii) the datasets $\mathcal{D}_{L} \leftarrow \mathcal{D}_{L} \cup \mathcal{S}_t$ and $\mathcal{D}_{U} \leftarrow \mathcal{D}_{U} \setminus \mathcal{S}_t$ are updated;
and, finally, (iv) the prompt vector $V$ is optimized by minimizing the cross-entropy loss over the current labeled set $\mathcal{D}_{L}$.

\subsection{Active Prompt Learning Framework} 
\label{sec:APL}

The recently proposed CB+SQ~\cite{kim2025active} efficiently adapt frozen VLMs to downstream tasks through three stages: class-guided clustering, cluster-balanced sampling, and selective querying.

\begin{figure}[htbp]
    \centering
    \includegraphics[width=0.9\linewidth]{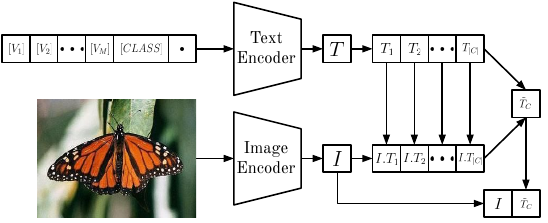}
    \caption{CB+SQ~\cite{kim2025active} class-guided features we used in our framework.}
    \label{fig:APL}
\end{figure}

For the class-guided clustering, the CLIP dual-encoder architecture is used to extract class-guided features, as shown in Figure~\ref{fig:APL}.
For a given image $x$, the image encoder $E_I$ computes an image feature $I$ and the text encoder $E_T$ computes a text feature $T$. 
The text feature is then weighted by the model's zero-shot similarity scores ($I.T$), generating $\tilde{T}_C$.

The dataset is then partitioned into $K$ distinct clusters by applying the $k$-means algorithm to the concatenation of the features $I$ and $\tilde{T}_C$.
To guarantee that the queried samples comprehensively cover the entire data distribution and avoid redundancy, the framework employs a cluster-balanced sampling strategy. The  most informative sample from each of the $k$ clusters are selected.
To identify this representative sample for a given cluster $C_i$, the framework calculates the centroid $c_i$ of each cluster by averaging the class-guided features of all samples in $C_i$.
The most informative sample is then defined as the image that lowest $L_2$ distance to the centroid.

Due to the strong zero-shot classification capabilities of the pre-trained VLM, the framework introduces a budget-saving selective querying step. Before a selected sample $x$ is sent for human annotation, its zero-shot prediction confidence is compared against an adaptive, class-wise threshold $\gamma_c$. If the sample's maximum predicted probability exceeds this threshold a pseudo-label $\hat{y}$ is assigned, avoiding the necessity of the human annotator.

\subsection{Unsupervised Transfer Clustering}
\label{sec:APL_TURTLE}

CB+SQ~\cite{kim2025active} rely on the $k$-means algorithm---a simplistic distance-based approach---for clustering and sample selection.
We propose to improve this strategy in our UTC+SQ framework by replacing the $k$-means clustering with unsupervised transfer techniques. These methods are capable of establishing semantically meaningful clusters by extracting high-fidelity pseudo-labels. 
To generate these semantic clusters, we explore two state-of-the-art unsupervised transfer frameworks:

\begin{itemize}
    \item \textbf{HUME~\cite{HUME}:} The HUME framework infers the underlying human labeling of a dataset without external supervision by evaluating the generalization error of linear models across different representation spaces.
    Two representation spaces are employed: the first corresponds to the task encoder and is acquired through task-specific self-supervised pre-training directly on the target dataset, while the second, associated with the linear model, is obtained from a frozen, large-scale pre-trained model.
    Specifically, the task encoder is parameterized as 
    $\tau_{\theta}^{HUME}(x) = \sigma(\theta^T \psi(x))$, where $\theta$ are the continuous parameters of the encoder, $\psi(x)$ is the representation space of the linear model, and $\sigma$ is an activation function.
    To evaluate and discover the optimal labeling, the following loss function is minimized over a test split $\mathcal{D}_{te}$:
    \begin{equation}
    \mathcal{L}^{HUME}(\theta) = \sum_{x \in \mathcal{D}_{te}} \mathcal{L}_{ce}(f_{approx}(x), \tau_{\theta}^{HUME}(x)).
    \end{equation}
    Here, $\mathcal{L}_{ce}$ represents the standard cross-entropy loss, $f_{approx}$ is an approximate solution to the optimal downstream linear classifier.
    Iterative differentiation is used to resolve this complex bilevel optimization problem. 

    \item \textbf{TURTLE~\cite{turtle}:} The TURTLE framework expands upon HUME by eliminating the need for task-specific representation learning. It searches for the dataset labeling that induces maximal-margin classifiers simultaneously across the representation spaces of $K$ VLM models.
    The task encoder is defined as an ensemble over $K$ representation spaces $\phi_k(x)$, formulated as $\tau_{\theta}^{TURTLE}(x) = \frac{1}{K} \sum_{k=1}^K \sigma(\theta_k^T \phi_k(x))$.
    The final cluster assignments, which act as our semantic pseudo-labels, are computed as $\arg\max_{c=1, ..., C} [\tau_{\theta}^{TURTLE}(x)]_c$.
    To achieve this goal, the following optimization objective is defined:
    \begin{align}
        \mathcal{L}_M^{TURTLE}(\theta) = \sum_{k=1}^K \sum_{x \in \mathcal{D}} \mathcal{L}_{ce}(w_M^k \phi_k(x); \tau_\theta(x)) \\
        \text{s.t. } w_M^k = \Xi^{(M)}(w_0^k, \mathcal{D}), \forall k \nonumber
    \end{align}
    where $\mathcal{D}$ is the dataset and $w_M^k$ are the weights of the $k$-th linear model trained to fit the labeling defined by $\tau_\theta$ using an iterative optimization algorithm $\Xi^{(M)}(w_0^k, \mathcal{D})$ that runs for $M$ steps starting from initial weights $w_0^k$.
\end{itemize}

Once the unlabeled pool $\mathcal{D}_U$ is partitioned into semantic clusters using the unsupervised transfer pseudo-labels, we must select the single most representative and informative sample from each cluster to send to the human oracle. We rigorously tested multiple selection strategies to define the optimal acquisition function:

\begin{itemize}
    \item \textbf{Centroids:} The feature centroid of the semantic cluster is calculated as an average of the samples in it. The sample with the smallest $L_2$ distance to the centroid is selected, ensuring representation of the cluster's core distribution.
    \item \textbf{Lowest Entropy:} The entropy of the task encoder's predicted probability distribution for each sample is calculated. The sample with the lowest entropy is selected, since it indicates the model is highly certain about its semantic cluster assignment.
    \item \textbf{Highest Confidence:} We select the sample that yields the maximum predicted probability for its assigned pseudo-label. This ensures the queried image represents the most prototypical instance of the semantic concept.
    \item \textbf{Margin Sampling:} This strategy selects the sample with the largest difference between its highest and second-highest predicted class probabilities. A large margin indicates strong discriminative certainty, avoiding samples that lie on the ambiguous decision boundaries.
    \item \textbf{Margin Sampling + Confidence:} A hybrid approach that weights both the margin sampling and highest confidence. This combination aims to select samples that are simultaneously highly representative of their own class and distinctly separated from the nearest competing class.
\end{itemize}

\section{Experimental Protocol}
\label{sec:experimental_protocol}

In this section we describe the datasets, implementation details, and evaluation metrics used.

\subsection{Datasets}

To evaluate our UTC+SQ framework, we followed previous works~\cite{HUME,turtle,kim2025active} and utilized the eight publicly available image classification datasets described in Table~\ref{tab:datasets}.

\begin{table}[htbp]
\centering
\setlength{\tabcolsep}{4.5pt}
\scriptsize
\caption{Description of the evaluated datasets.}
\label{tab:datasets}
    \begin{tabular}{lcccccc}
    \toprule
    \textbf{Dataset} & \textbf{Classes} & \textbf{Train} & \textbf{Validation} & \textbf{Test} & \textbf{Resolution} \\
    \midrule
    CIFAR-10   & 10  & 45,000 & 5,000 & 10,000 & 32$\times$32 \\
    STL-10     & 10  & 4,500  & 500   & 8,000  & 96$\times$96 \\
    Caltech101 & 100 & 4,128  & 1,649 & 2,465  & $\sim$300$\times$200 \\
    Flowers102 & 102 & 4,093  & 1,633 & 2,463  & 500 px (smaller side) \\
    OxfordPets & 37  & 2,944  & 736   & 3,669  & Multiple \\
    Aircraft   & 100 & 3,334  & 3,333 & 3,333  & 1--2 MP \\
    DTD        & 47  & 2,820  & 1,128 & 1,692  & 300$\times$300 -- 640$\times$640 \\
    EuroSAT    & 10  & 13,500 & 5,400 & 8,100  & 64$\times$64 \\
    \bottomrule
    \end{tabular}
\end{table}

All images are resized to the ViT-B/16 standard input resolution of 224$\times$224 pixels.
For CIFAR-10, STL-10, we followed the standard data splits of the original datasets, while for the remainder datasets, we used the same configurations as CB+SQ~\cite{kim2025active}.
These datasets span diverse domains, including general object recognition, fine-grained classification, and specialized textures, ensuring a rigorous benchmark. 

\subsection{Implementation Details}
\label{sec:implementation_details}

For our VLM backbone, we employ the pre-trained CLIP architecture with a ViT-B/16 vision encoder. During the APL phase, the weights of both the image and text encoders remain frozen to preserve prior knowledge.

The prompt parameters are updated using the SGD optimizer with a momentum of 0.9 and an initial learning rate of 0.002. The learning rate decays over time according to a cosine annealing scheduler over 200 epochs per active learning round, utilizing a batch size of 32. Furthermore, no description augmentations were used, and we kept the same active learning method as CB+SQ~\cite{kim2025active}.

For the unsupervised transfer frameworks, HUME used image features from DINOv2 (ViT-B/16) and CB+SQ~\cite{kim2025active} class-guided features as the two feature spaces. Notably, HUME was trained for 200 iterations instead of the original 1,000, since the accuracy stabilized after fewer iterations.

For TURTLE, we also adopted the class-guided features~\cite{kim2025active}. Here, the learning rates were set to default values, namely 0.005 for the task encoder and 0.001 for the linear model. Additionally, the task encoder was adjusted at each round according to the variation in the number of clusters.

The initial annotation budget is set to the number of target classes, $B_0 = C$.
For each subsequent round $r$, the budget is incremented, and the number of clusters $K$ increases linearly according to $K = C \times r$.
All other hyperparameters and protocols match the original works.

Experiments were executed on a machine with an Intel Core(TM) i9-10900X 10-core CPU, 64 GB of DDR4 memory and a NVIDIA RTX 2080 Ti GPU. The system runs Ubuntu 22.04.4 LTS with Linux kernel 5.15.0-69 and ext4 file system.

\subsection{Evaluation Metrics}
\label{sec:metrics}

APL is inherently sensitive to the specific samples chosen during the initial query.
To ensure that our performance improvements are statistically robust, all experiments are executed across three independent runs. We report the mean classification accuracy of these runs alongside the standard deviation on the test set. A lower standard deviation coupled with higher accuracy indicates that an approach consistently finds representative samples, effectively mitigating the high variance typically associated with the cold-start problem.
Performance metrics are reported both per dataset and as an overall average. The overall average accuracy and its corresponding standard deviation are calculated over the average performance across all datasets from each run.

\begin{table*}[!htb]
\centering
\scriptsize
\caption{Ground-truth alignment (clustering accuracy) for $k$-means, HUME and TURTLE. Results are reported as Mean Accuracy (\%) $\pm$ Standard Deviation. The best performance for each dataset is highlighted in bold, and the second best is underlined.}
\label{tab:clustering}
\begin{tabular}{lccccccccc}
\toprule
\textbf{Method} & \textbf{Flowers102} & \textbf{OxfordPets} & \textbf{DTD} & \textbf{Caltech101} & \textbf{Aircraft} & \textbf{EuroSAT} & \textbf{CIFAR-10} & \textbf{STL-10} & \textbf{Average} \\
\midrule
\textbf{$K$-Means~\cite{kim2025active}} & \textbf{78.92 $\pm$ 2.31} & \underline{85.22 $\pm$ 3.59} & \underline{51.97 $\pm$ 1.92} & \textbf{71.87 $\pm$ 3.16} & \underline{30.69 $\pm$ 0.51} & 65.41 $\pm$ 4.54 & 78.60 $\pm$ 0.51 & 85.11 $\pm$ 1.80 & \underline{68.47 $\pm$ 0.92} \\
\textbf{HUME~\cite{HUME}} & 67.85 $\pm$ 0.98 & 82.59 $\pm$ 3.20 & 51.48 $\pm$ 0.82 & 55.85 $\pm$ 0.79 & 20.43 $\pm$ 0.58 & \underline{66.33 $\pm$ 3.34} & \textbf{96.65 $\pm$ 0.17} & \underline{94.71 $\pm$ 8.01} & 66.99
$\pm$ 1.11 \\
\textbf{TURTLE~\cite{turtle}} & \underline{72.92 $\pm$ 0.14} & \textbf{91.97 $\pm$ 0.06} & \textbf{57.23 $\pm$ 0.96} & \underline{59.34 $\pm$ 0.65} & \textbf{32.49 $\pm$ 0.08} & \textbf{77.47 $\pm$ 0.28} & \underline{93.18 $\pm$ 0.08} & \textbf{98.32 $\pm$ 0.00} & \textbf{72.86 $\pm$ 0.19} \\
\bottomrule
\end{tabular}
\end{table*}

\begin{table*}[!htb]
\centering
\setlength{\tabcolsep}{4.5pt}
\scriptsize
\caption{Comparison of CB+SQ~\cite{kim2025active} against ours UTC+SQ framework variants in the first round of APL. Results are reported as mean accuracy (\%) $\pm$ standard deviation.}
\begin{tabular}{lccccccccc}
\toprule
\textbf{Method / Strategy} & \textbf{Flowers102} & \textbf{OxfordPets} & \textbf{DTD} & \textbf{Caltech101} & \textbf{Aircraft} & \textbf{EuroSAT} & \textbf{CIFAR-10} & \textbf{STL-10} & \textbf{Average} \\
\midrule
\multicolumn{9}{l}{\textbf{Baseline}} \\
\textbf{CB+SQ~\cite{kim2025active}} & \textbf{80.05 $\pm$ 0.72} & 78.87 $\pm$ 4.20 & 43.72 $\pm$ 1.69 & \textbf{83.38 $\pm$ 1.88} & \textbf{24.84 $\pm$ 0.47} & 51.94 $\pm$ 4.72 & 65.58 $\pm$ 0.79 & 78.66 $\pm$ 1.78 & 63.38 $\pm$ 0.39 \\
\midrule
\multicolumn{10}{l}{\textbf{UTC+SQ w/ HUME (ours)}} \\
\textbf{Centroids} & 78.93 $\pm$ 0.18 & 79.33 $\pm$ 1.03 & 43.05 $\pm$ 1.01 & 78.16 $\pm$ 1.20 & 19.58 $\pm$ 0.77 & \textbf{57.94 $\pm$ 7.05} & \textbf{80.33 $\pm$ 1.14} & 85.50 $\pm$ 1.17 & 65.35 $\pm$ 0.66\\
\textbf{Lowest Entropy} & 69.85 $\pm$ 1.18 & 70.58 $\pm$ 3.72 & 39.36 $\pm$ 5.90 & 71.74 $\pm$ 0.49 & 17.07 $\pm$ 2.30 & 50.91 $\pm$ 1.93 & 71.06 $\pm$ 4.28 & 84.93 $\pm$ 6.32 & 59.44 $\pm$ 1.09 \\
\textbf{Highest Confidence} & 73.54 $\pm$ 2.77 & 76.47 $\pm$ 2.10 & 42.55 $\pm$ 0.53 & 75.20 $\pm$ 1.00 & 20.37 $\pm$ 0.52 & 53.44 $\pm$ 5.35 & 74.13 $\pm$ 3.80 & 84.86 $\pm$ 7.12 & 62.57 $\pm$ 1.37 \\
\textbf{Margin Sampling} & 74.42 $\pm$ 1.72 & 73.14 $\pm$ 2.51 & 42.26 $\pm$ 2.61 & 75.55 $\pm$ 1.78 & 19.73 $\pm$ 0.65 & 49.91 $\pm$ 2.10 & 71.33 $\pm$ 4.62 & 82.28 $\pm$ 6.08 & 61.08 $\pm$ 1.84 \\
\textbf{Margin Samp. + Conf.} & 74.66 $\pm$ 1.70 & 73.84 $\pm$ 3.31 & 42.51 $\pm$ 1.49 & 76.04 $\pm$ 2.34 & 20.11 $\pm$ 0.96 & 50.06 $\pm$ 2.40 & 70.89 $\pm$ 2.56 & 83.74 $\pm$ 4.77 & 61.48 $\pm$ 1.38 \\
\midrule
\multicolumn{10}{l}{\textbf{UTC+SQ w/ TURTLE (ours)}} \\
\textbf{Centroids} & \underline{80.03 $\pm$ 0.88} & \textbf{83.85 $\pm$ 0.83} & 45.15 $\pm$ 1.11 & \underline{78.81 $\pm$ 0.82} & 24.03 $\pm$ 0.58 & 54.42 $\pm$ 3.60 & 72.99 $\pm$ 2.07 & 83.95 $\pm$ 4.59 & 65.40 $\pm$ 0.91 \\
\textbf{Lowest Entropy} & 77.56 $\pm$ 1.04 & 80.87 $\pm$ 0.95 & \underline{46.51 $\pm$ 0.91} & 78.12 $\pm$ 1.06 & 24.17 $\pm$ 0.74 & 54.56 $\pm$ 2.47 & 75.94 $\pm$ 2.33 & \textbf{87.66 $\pm$ 5.24} & \underline{65.67 $\pm$ 1.08} \\
\textbf{Highest Confidence} & 77.73 $\pm$ 0.38 & 80.35 $\pm$ 1.68 & 44.94 $\pm$ 0.59 & 78.73 $\pm$ 1.44 & \underline{24.38 $\pm$ 1.38} & \underline{55.91 $\pm$ 2.24} & \underline{77.38 $\pm$ 1.17} & \textbf{87.66 $\pm$ 5.24} & \textbf{65.89 $\pm$ 0.90} \\
\textbf{Margin Sampling} & 77.35 $\pm$ 0.51 & \underline{80.95 $\pm$ 1.09} & 46.22 $\pm$ 0.10 & 77.96 $\pm$ 0.95 & 24.15 $\pm$ 0.53 & 54.93 $\pm$ 1.41 & 75.72 $\pm$ 0.81 & \underline{87.57 $\pm$ 5.16} & 65.60 $\pm$ 0.74 \\
\textbf{Margin Samp. + Conf.} & 77.48 $\pm$ 0.74 & 80.32 $\pm$ 1.64 & \textbf{46.92 $\pm$ 2.07} & 78.23 $\pm$ 1.45 & 23.97 $\pm$ 0.44 & 54.93 $\pm$ 1.41 & 75.56 $\pm$ 1.01 & \textbf{87.66 $\pm$ 5.24} & 65.63 $\pm$ 0.81 \\
\bottomrule
\end{tabular}
\label{tab:round1_ablation}
\vspace{-1mm}
\end{table*}

\section{Experimental Results}
\label{sec:experimental_results}

To isolate the contributions of our UTC+SQ framework, we first evaluate the quality of the initial clusters formed prior to any human querying.
Table~\ref{tab:clustering} presents the ground-truth alignment (clustering accuracy) on the test set for the simplistic distance-based clustering method ($k$-means) against unsupervised transfer learning models (HUME and TURTLE).
The results demonstrate that replacing the $k$-means clustering with unsupervised transfer yields substantial improvements. Specifically, TURTLE emerges as the most robust strategy, achieving the highest average clustering accuracy.
Conversely, HUME obtained the best results only for CIFAR-10, struggling in the other datasets.
The $k$-means strategy remained the best only on the Flowers102 and Caltech101 datasets, but it suffered from high variance and poor alignment on datasets such as EuroSAT and STL-10.
These findings validate our hypothesis that unsupervised transfer can establish highly representative semantic clusters.

Next, we evaluate the impact of our framework in mitigating the cold-start problem during the critical first round of APL. Table~\ref{tab:round1_ablation} compares the baseline CB+SQ~\cite{kim2025active} (which relies on the $k$-means algorithm) against different sample selection strategies applied to our UTC+SQ framework with the HUME and TURTLE unsupervised transfer methods.

While CB+SQ performs strongly on specific datasets with well-separated and cohesive features, achieving top accuracy on Flowers102, Caltech101, and Aircraft, it struggles with consistency across diverse domains, where its simplistic distance-based approach fails to handle feature spaces with complex distributions and poorly separated clusters.
Conversely, UTC+SQ with TURTLE when paired with appropriate selection strategies was able to outperform this baseline. 

When comparing selection strategies for our method, Centroids obtained good results with TURTLE on datasets with well-separated clusters, like Flowers102 and Caltech101. 
However, on more challenging datasets featuring clusters without clear separability, such as CIFAR-10 and EuroSAT, Highest Confidence performed better by selecting more representative examples.
We selected UTC+SQ with TURTLE (Highest Confidence) as our best method, since it obtained the highest average accuracy (65.89\%), being able to outperform CB+SQ on 5 out of 8 datasets: OxfordPets (80.35\%), DTD (46.92\%), EuroSAT (55.91\%), CIFAR-10 (77.38\%), and STL-10 (87.66\%).
Our UTC+SQ with HUME (Centroids) also showed isolated success, particularly on EuroSAT (57.94\%) and CIFAR-10 (80.33\%).
Overall, replacing spatial distance-based clustering with TURTLE's semantic pseudo-labels, combined with Highest Confidence sampling, ensures that the initial query budget is utilized with maximum efficiency.

\begin{figure*}[!htb]
    \centering
    \includegraphics[width=0.9\linewidth]{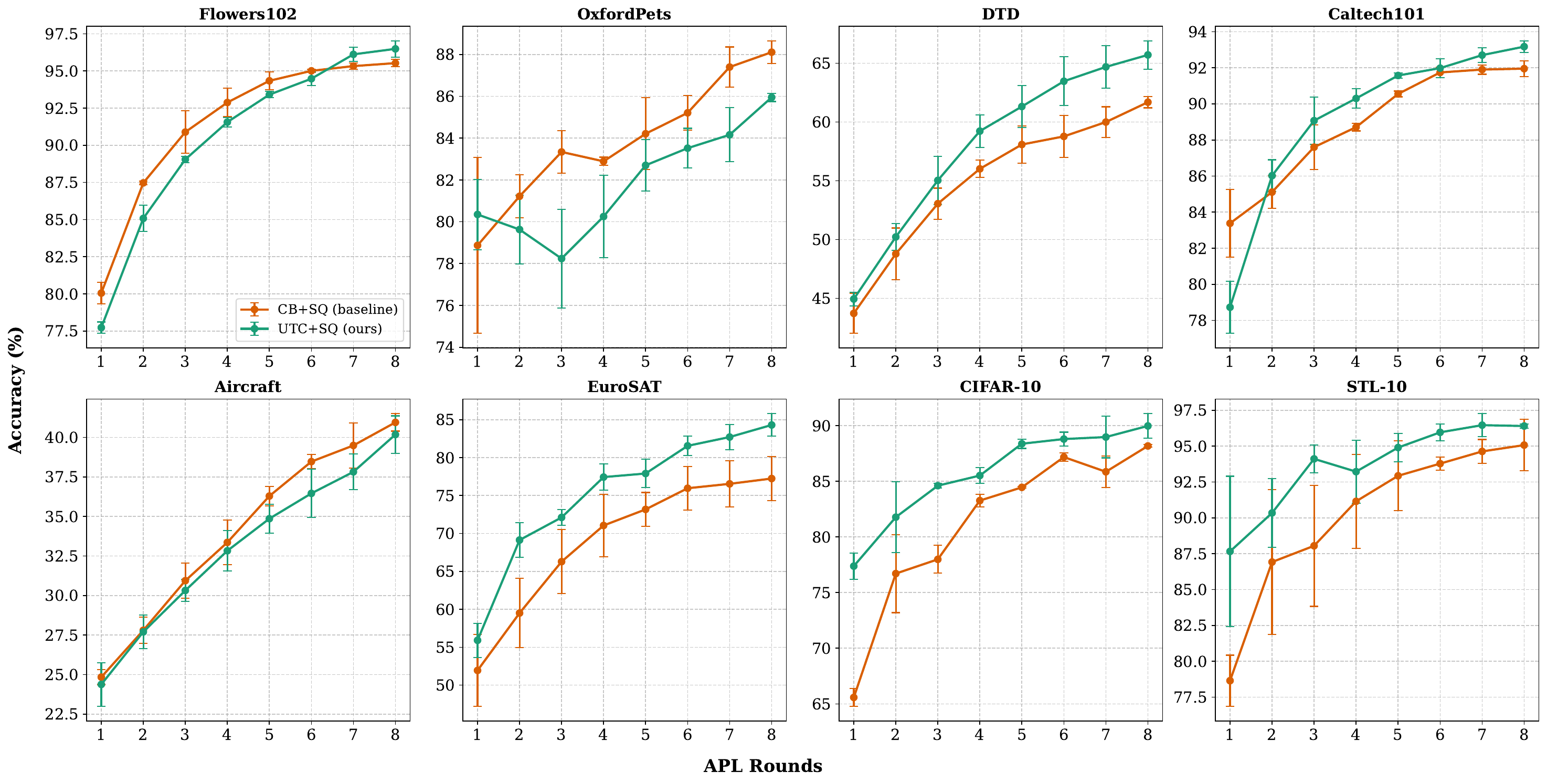}
    \caption{Mean accuracy and standard deviation comparing the baseline CB+SQ~\cite{kim2025active} against the best configuration of UTC+SQ framework across 8 APL rounds.}
    \label{fig:final_results}
\end{figure*}

Owing to its strong cold-start performance, our method is particularly well suited to real-world scenarios with limited annotation budgets, where AL can be performed in only a few—or even a single—round.
Since mitigating the cold-start problem benefits all subsequent rounds, we evaluated the best configuration of our UTC+SQ framework against the CB+SQ baseline across 8 full rounds of APL. Figure~\ref{fig:final_results} illustrates the learning curves, comparing the mean accuracy and standard deviation for both methods across the eight evaluated datasets.

On DTD, EuroSAT, CIFAR-10, and STL-10, our UTC+SQ framework established a strong initial advantage over the baseline that expanded across all rounds (e.g., reaching 96.40\% vs. 95.07\% on STL-10 at the last round).
Conversely, on Caltech101 and Flowers102, the baseline initially outperformed our method.
However, the superior representativeness of the semantic clusters allowed our method to close the gap and surpass the baseline by round 8 (e.g., 93.17\% vs. 91.95\% on Caltech101).
On the fine-grained Aircraft dataset, both methods exhibited similar performance, while on OxfordPets, the baseline recovered from a round 1 deficit to finish slightly ahead (88.11\% vs. 85.95\%). Overall, the mean accuracy trajectories across the rounds confirm that our UTC+SQ framework---which replaces $k$-means with unsupervised transfer---provides a highly effective foundation, achieving the highest final accuracy on 6 out of the 8 benchmarks.

Beyond effectiveness, efficiency is also crucial for AL. Although TURTLE solves a complex margin-maximization optimization problem, it is surprisingly efficient (see Figure~3 of~\cite{turtle}). While $k$-means is faster on smaller datasets, its execution time increases as the number of samples and classes grow, whereas TURTLE remains nearly constant~\cite{turtle}. Therefore, our method is not only effective but also efficient and scalable.

\section{Conclusions}
\label{sec:concluions}

In this paper, we addressed the cold-start problem inherent in APL frameworks. Traditional approaches, such as CB+SQ~\cite{kim2025active}, rely on $k$-means clustering for sample selection, which limits their ability to capture complex data distributions. To improve this approach, we proposed the UTC+SQ framework, which leverages state-of-the-art unsupervised transfer learning techniques to partition the unlabeled data pool into high-fidelity, semantically meaningful clusters.

Our comprehensive evaluations across eight diverse image classification datasets demonstrated that transitioning from distance-based to projection-based clustering improves the representativeness of the initial queried subset in the majority of datasets.
By evaluating multiple unsupervised transfer models and selection strategies, we identified TURTLE~\cite{turtle} as the best approach for clustering and that prioritizing the samples with the highest confidence maximizes prompt-tuning performance.
Consequently, our UTC+SQ framework successfully mitigates the initial selection bias and consistently outperforms the current state-of-the-art APL baseline in mean test accuracy.

Future research will explore the integration of these unsupervised semantic clusters with open-set recognition strategies to handle unconstrained environments with unknown classes.

\section*{Acknowledgment}

\iffinal

This research was supported by FAPESP (\#2023/17577-0, \#2024/04500-2, \#2025/15222-6) and CNPq (\#315220/2023-6, \#420442/2023-5, \#444982/2024-8, \#164065/2025-3).

\else

The authors would like to thank...

\fi

\bibliographystyle{IEEEtran}
\bibliography{refs.bib}

@article{hume,
  title={The pursuit of human labeling: a new perspective on unsupervised learning},
  author={Gadetsky, Artyom and Brbic, Maria},
  journal={NeurIPS},
  volume={36},
  pages={60527--60546},
  year={2023}
}

@article{turtle,
    author = {Gadetsky, Artyom and Jiang, Yulun and Brbi\'{c}, Maria},
    title = {Let go of your labels with unsupervised transfer},
    year = {2024},
    publisher = {JMLR.org},
    journal = {ICML},
    articleno = {575},
    numpages = {26},
}

@article{
    kim2025active,
    title={Active Prompt Learning with Vision-Language Model Priors},
    author={Hoyoung Kim and Seokhee Jin and Changhwan Sung and Jaechang Kim and Jungseul Ok},
    journal={TMLR},
    publisher={JMLR Inc.},
    issn={2835-8856},
    year={2025}
}

@article{bang2024active,
  title={Active prompt learning in vision language models},
  author={Bang, Jihwan and Ahn, Sumyeong and Lee, Jae-Gil},
  journal={CVPR},
  publisher={IEEE},
  pages={27004--27014},
  year={2024}
}

@article{chen2022making,
  title={Making Your First Choice: To Address Cold Start Problem in Vision Active Learning},
  author={Chen, Liangyu and Bai, Yutong and Huang, Siyu and Lu, Yongyi and Wen, Bihan and Yuille, Alan L and Zhou, Zongwei},
  journal={CoRR, abs/2210.02442},
  year={2022}
}

@article{radford2021learning,
  title={Learning transferable visual models from natural language supervision},
  author={Radford, Alec and Kim, Jong Wook and Hallacy, Chris and Ramesh, Aditya and Goh, Gabriel and Agarwal, Sandhini and Sastry, Girish and Askell, Amanda and Mishkin, Pamela and Clark, Jack and others},
  journal={ICML},
  pages={8748--8763},
  year={2021},
  organization={PmLR}
}

@article{gupte2024revisiting,
title={Revisiting Active Learning in the Era of Vision Foundation Models},
author={Sanket Rajan Gupte and Josiah Aklilu and Jeffrey J Nirschl and Serena Yeung-Levy},
journal={TMLR},
publisher={JMLR Inc.},
issn={2835-8856},
year={2024},
note={}
}

@article{bayer2026activellm,
  title={Activellm: Large language model-based active learning for textual few-shot scenarios},
  author={Bayer, Markus and Lutz, Justin and Reuter, Christian},
  journal={Transactions of the Association for Computational Linguistics},
  volume={14},
  pages={1--22},
  year={2026},
  publisher={MIT Press}
}

@article{achiam2023gpt,
  title={Gpt-4 technical report},
  author={Achiam, Josh and Adler, Steven and Agarwal, Sandhini and Ahmad, Lama and Akkaya, Ilge and Aleman, Florencia Leoni and Almeida, Diogo and Altenschmidt, Janko and Altman, Sam and Anadkat, Shyamal and others},
  journal={CoRR, abs/2303.08774},
  year={2023}
}

@article{zhou2022learning,
  title={Learning to prompt for vision-language models},
  author={Zhou, Kaiyang and Yang, Jingkang and Loy, Chen Change and Liu, Ziwei},
  journal={IJCV},
  volume={130},
  number={9},
  pages={2337--2348},
  year={2022},
  publisher={Springer}
}

@article{khattak2023maple,
  title={Maple: Multi-modal prompt learning},
  author={Khattak, Muhammad Uzair and Rasheed, Hanoona and Maaz, Muhammad and Khan, Salman and Khan, Fahad Shahbaz},
  journal={CVPR},
  publisher={IEEE},
  pages={19113--19122},
  year={2023}
}

@article{zhu2023prompt,
  title={Prompt-aligned gradient for prompt tuning},
  author={Zhu, Beier and Niu, Yulei and Han, Yucheng and Wu, Yue and Zhang, Hanwang},
  journal={ICCV},
  publisher={IEEE},
  pages={15659--15669},
  year={2023}
}

@article{saltori2022lowbudget,
  author       = {Cristiano Saltori and
                  Paolo Rota and
                  Nicu Sebe and
                  Jurandy Almeida},
  title        = {Low-budget label query through domain alignment enforcement},
  journal      = {CVIU},
  volume       = {222},
  pages        = {103485},
  year         = {2022}
}

\end{document}